\documentclass{article}

\usepackage[numbers,sort&compress]{natbib}
\usepackage[preprint]{neurips_2024}

\usepackage[utf8]{inputenc} 
\usepackage[T1]{fontenc}    
\usepackage[pagebackref=true,breaklinks=true,colorlinks,bookmarks=false]{hyperref}     
\usepackage{url}            
\usepackage{booktabs}       
\usepackage{amsfonts}       
\usepackage{nicefrac}       
\usepackage{microtype}      
\usepackage{xcolor}         
\usepackage{multirow}
\usepackage{graphicx}
\usepackage{amsmath}
\usepackage{enumitem}
\usepackage{subfig}
\usepackage{dsfont}

\usepackage[capitalise]{cleveref}

\crefname{table}{Tab.}{Tab.}
\crefname{section}{Sec.}{Sec.}

\def\Approach{EditScout}
\def\Ourtestset{PerfBrush}

\def\1n{\mathbf{1}_n}
\def\0{\mathbf{0}}
\def\1{\mathbf{1}}

\definecolor{pink}{rgb}{0.9,0.5,0.5}
\definecolor{purple}{rgb}{0.5, 0.4, 0.8}   
\definecolor{gray}{rgb}{0.3, 0.3, 0.3}
\definecolor{mygreen}{rgb}{0.2, 0.6, 0.2}

\definecolor{greena}{rgb}{0.4, 0.5, 0.1}

\definecolor{bluea}{rgb}{0, 0.4, 0.6}

\definecolor{reda}{rgb}{0.6, 0.2, 0.1}

\newcommand{\cm}[1]{}

\newcommand{\myheading}[1]{\vspace{1ex}\noindent \textbf{#1}}

\newif\ifshowsolution
\showsolutiontrue

\ifshowsolution

\else

\fi

\definecolor{mydarkblue}{rgb}{0,0.08,1}
\definecolor{mycolor2}{rgb}{1.0,1.0,0.0}
\definecolor{mycolor1}{rgb}{0,0.5,1}

\definecolor{mydarkgreen}{rgb}{0.02,0.6,0.02}
\definecolor{myred}{rgb}{1.0,0.0,0.0}
\definecolor{mypurple}{rgb}{111,0,255}

\usepackage{soul}

\usepackage{amssymb}
\usepackage{pifont}
\title{\Approach: Locating Forged Regions from Diffusion-based Edited Images with Multimodal LLM}

\author{%
  Quang Nguyen$^{1}$ \quad Truong Vu$^{1}$ \quad Trong-Tung Nguyen$^{1}$ \quad  Yuxin Wen$^{2}$
  \\ 
  \textbf{Preston K Robinette}$^{3}$ \quad \textbf{Taylor T Johnson}$^{3}$ \quad \textbf{Tom Goldstein}$^{2}$ \quad \textbf{Anh Tran}$^{1}$ \quad \textbf{Khoi Nguyen}$^{1}$ \\ \\ 
  $^1$VinAI Research  \quad
  $^2$ University of Maryland  \quad $^3$Vanderbilt University
}

\begin{document}
\maketitle


\begin{abstract}
\label{sec:abstract}

Image editing technologies are tools used to transform, adjust, remove, or otherwise alter images. Recent research has significantly improved the capabilities of image editing tools, enabling the creation of photorealistic and semantically informed forged regions that are nearly indistinguishable from authentic imagery, presenting new challenges in digital forensics and media credibility. While current image forensic techniques are adept at localizing forged regions produced by traditional image manipulation methods, current capabilities struggle to localize regions created by diffusion-based techniques. To bridge this gap, we present a novel framework that integrates a multimodal Large Language Model (LLM) for enhanced reasoning capabilities to localize tampered regions in images produced by diffusion model-based editing methods. By leveraging the contextual and semantic strengths of LLMs, our framework achieves promising results on MagicBrush, AutoSplice, and PerfBrush (novel diffusion-based dataset) datasets, outperforming previous approaches in mIoU and F1-score metrics.  Notably, our method excels on the PerfBrush dataset, a self-constructed test set featuring previously unseen types of edits. Here, where traditional methods typically falter, achieving markedly low scores, our approach demonstrates promising performance.


\end{abstract}
\section{Introduction}
\label{sec:introduction}

In recent years, the landscape of image editing has seen substantial advancement as powerful editing tools have expanded beyond basic copy-move and splicing techniques, embracing more sophisticated modifications. Particularly notable are diffusion-based editing methods \cite{ramesh2021zero, ju2024brushnet, yang2023magicremover, 
avrahami2022blended, nguyen2024flex}, which enable the creation of photorealistic forged regions seamlessly integrated with the style and lighting of the surroundings, posing challenges for both human perception and machine forensic techniques in localizing these alterations. While these tools offer immense creative potential, their misuse carries significant risks.

\begin{figure}[!t]
    \centering
    \includegraphics[width=1.\linewidth]{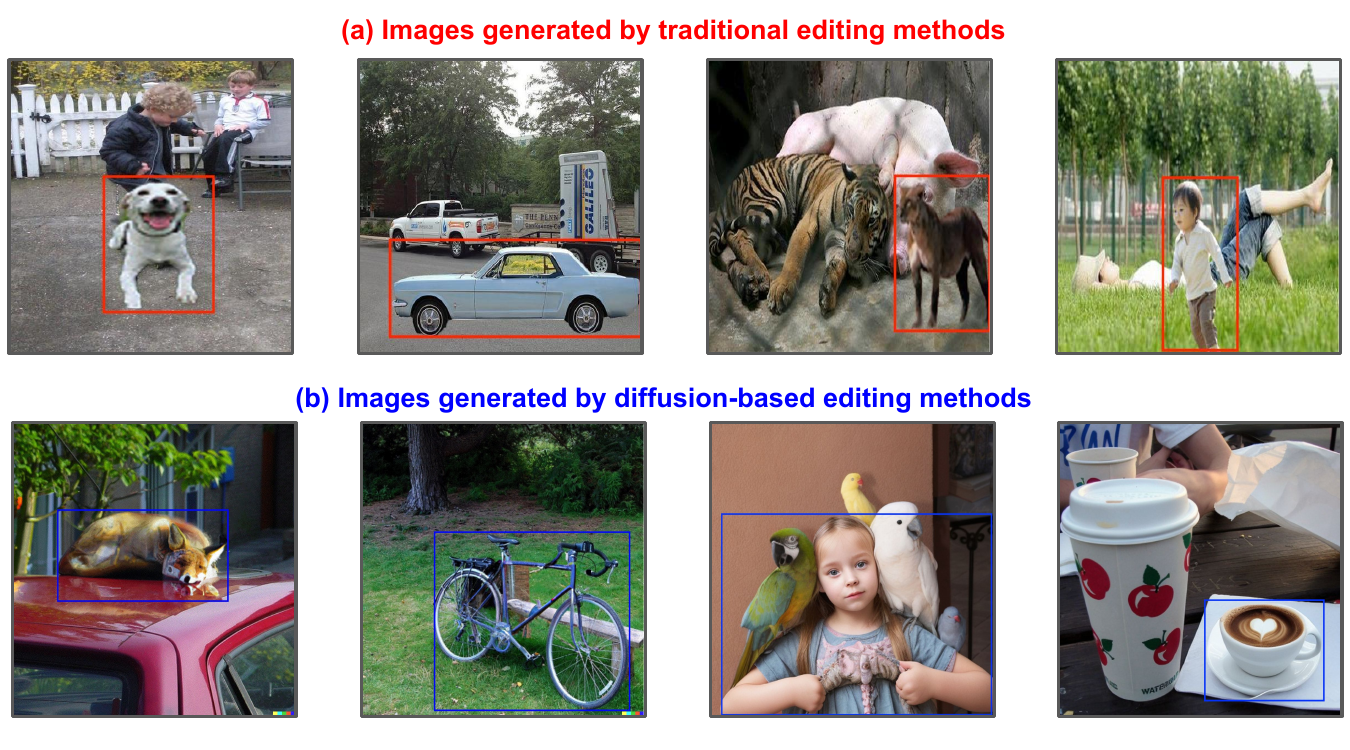}
    \caption{Visualizing edited images with annotated bounding boxes highlights notable disparities between (a) traditional editing methods (\textcolor{red}{boxes in 
 red}) and (b) diffusion-based techniques (\textcolor{blue}{boxes in 
 blue}). While conventional edits are easily identifiable, diffusion-based alterations present challenges in image forensics due to their seamless blending of edited regions with their surroundings, yielding photorealistic results.}
    \label{fig:compare_trad_diff}
\end{figure}

Consequently, traditional image forensic methods struggle to localize the forged regions created by diffusion-based editing techniques compared to those produced by traditional image manipulation methods. One hypothesis is that the highly realistic nature of images edited by diffusion models enables them to blend seamlessly with the original, making it difficult to detect using traditional approaches that rely on detecting internal inconsistencies and editing traces of edited images as illustrated in ~\cref{fig:compare_trad_diff}. This indicates a pressing need for new approaches in image forensics that can adapt to the complexities introduced by advanced diffusion-based editing tools.

Hence, this paper is dedicated to tackling the intricate challenges posed by diffusion-based image editing techniques, which stand at the forefront of the evolving field of image forensics. Specifically, \textit{given a suspicious image that appears to have been edited, our goal is to segment and identify the forged regions}. Next, we present \Approach, a novel methodology leveraging Multimodal Large Language Models (MLLMs) to precisely locate forgeries in diffusion-based edited images. 

This approach consists of two core modules: MLLM-based reasoning query generation and a segmentation network. The MLLM utilizes visual features extracted from a pretrained visual encoder like CLIP \cite{radford2021learning} along with a prompt for localizing edited regions to generate the reasoning query. Subsequently, this query is fed into the SAM to produce the segmentation mask. By integrating MLLMs, our method harnesses rich multimodal data, combining visual and linguistic cues to conduct a thorough, systematic analysis of images. The key advantage of employing MLLMs lies in their innate capacity to interpret and synthesize diverse data streams, enabling them to discern subtle high-level features and extract underlying semantic meanings indicative of tampering or manipulation.


To assess the effectiveness of \Approach~across various scenarios, including both familiar and novel editing types, we conduct training on the MagicBrush \cite{zhang2024magicbrush} and Autosplice \cite{jia2023autosplice} datasets and evaluate on multiple test sets. These include the test set from MagicBrush (representing seen editing types), CocoGLIDE \cite{dai2024instructblip} (exhibiting similar editing types), and our newly introduced test set, \Ourtestset~(featuring unseen editing types). Our method, \Approach, surpasses traditional image forensic techniques in performance across these datasets, particularly excelling in identifying unseen editing types. This underscores the robustness and generalizability of our proposed approach.

In summary, the contributions of this work are the following:
\begin{itemize} 

    \item We introduce a novel approach to image forensics, leveraging an MLLM for locating diffusion-based edits. This study represents the first attempt to explore MLLM capabilities in image forensic localization.
    \item We present \Ourtestset, a new image forensics test set specifically designed for detecting forged regions in diffusion-based edited images.
\end{itemize}

\section{Related Work}
\label{sec:related-work}

\myheading{Multimodal Large Language Models (MLLM).} With the advanced development of large language models such as GPT-3 \citep{brown2020language-gpt3}, Llama \citep{touvron2023llama}, Mistral \citep{jiang2023mistral}, Flan-T5 \citep{chung2024scaling-flant5}, Vicuna \citep{zheng2024judging-vicuna}, and PaLM \citep{chowdhery2023palm}, there is an increasing effort to apply these models for multimodal tasks, including the understanding and incorporation of visual information. Visual instruction tuning, which involves fine-tuning LLMs with visual supervision, has garnered significant interest and delivered surprising results. Several models have been developed using this method, including LLaVA \citep{liu2024visual}, MiniGPT-4 \citep{zhu2023minigpt}, InstructBLIP \citep{dai2024instructblip}, and other models \citep{gong2023multimodal, ye2023mplug, li2023mimic}. These MLLMs have demonstrated promising results in visual perception tasks such as detection \citep{ma2024groma, you2023ferret, Pi2023-detGPT, Zhang2023-nextchat, Li2023-covlm} and segmentation \citep{Lai2023-lisa, Yang2023-lisa++, Yuan2023-opsrey, Zhang2024-groundhog, Wu2023-see_say_segment, Ren2023-pixelLM, Zhang2023-nextchat}. Furthermore, they have recently been studied for their ability to detect generated images and deepfakes  \citep{fakebenchLi2024}. However, the application of such models in localizing tampering through editing forgery remains unexplored. Therefore, in this paper, we aim to study the potential capability of MLLMs in localizing diffusion-based editing forgery.

\myheading{MLLMs for image segmentation.}
MLLMs have significantly improved segmentation capability by integrating reasoning and deeper regional understanding. Several works such as LISA \citep{Lai2023-lisa} and CoRes \citep{bao2024cores} focus on reasoning-based segmentation, enhancing the ability to interpret intricate queries and generate acceptable segmentation masks. PixelLM \citep{Ren2023-pixelLM} and GROUNDHOG \citep{Zhang2024-groundhog}, which use efficient decoders and feature extractors to handle multi-target tasks and reduce object hallucination, hence improving the performance of segmentation tasks. Whereas, DeiSAM \citep{shindo2024deisam} integrates logic reasoning with multimodal large language models to improve referring expression segmentation. See, Say, Segment \citep{Wu2023-see_say_segment} employs joint training to maintain performance across tasks and prevent catastrophic forgetting. These studies inspires that with enhanced reasoning capabilities and regional understanding, MLLMs might offer an effective and explainable solution for forgery localization.

\myheading{Diffusion-based editing.}
Recent advancements in diffusion-based editing  methods have significantly enhanced the precision and realism of filling missing regions in images. Traditional techniques, which relied on hand-crafted features, often struggle with control over the inpainted content. However, the introduction of text-to-image diffusion models like Stable Diffusion \citep{rombach2022high-ldm} and Imagen \citep{saharia2022photorealistic-imagen} has enabled more controlled editing by using multimodal conditions such as text descriptions and segmentation maps. Models such as GLIDE \citep{nichol2021glide} and Stable Inpainting have further refined these pretrained models for editing tasks, employing random masks to effectively utilize context from unmasked regions. Additionally, techniques like Blended Diffusion \citep{avrahami2022blended} and Blended Latent Diffusion \citep{avrahami2023blended} use pretrained models and CLIP-guided reconstruction to improve mask-based editting quality. Innovative approaches such as Inpaint Anything \citep{yu2023inpaint} and MagicRemover \citep{yang2023magicremover} combine various models and user-friendly interfaces to offer greater flexibility and precision in image editing. These methods yield highly realistic results that are often indistinguishable from the original content, presenting substantial challenges for detection and localization.

\myheading{Image forgery localization} is a critical aspect of digital image forensics that focuses on pixel-level binary classification within an image. Most existing studies employ pixel-wise or patch-based \citep{mayer2018learned_patch} approaches to identify forged regions. Several methods combine image and noise features for more accurate forgery localization \citep{eitl@2024, Guillaro2022-trufor}. For instance, TruFor \citep{Guillaro2022-trufor} extract the high-level and low-level traces through transformer-based architecture that combines RGB image and noise-sensitive fingerprint. CAT-Net utilize the JPEG compression artifacts during image editing for localize the forgerd regions. Furthermore, HiFi \citep{Guo2023-hifi} proposed a hierachical approach for detecting and localizing the tampering regions based on the frequency domain. Whereas, PSCC-Net \citep{Liu2021-pscc} utilizes a progressive spatial-correlation module that leverages multi-scale features and cross-connections to generate binary masks in a coarse-to-fine manner. 
While these methods yield good results for various forgery types, such as copy-move and splicing, they cannot deliver an acceptable performance when faced with diffusion-based inpainting methods. The high realism and indistinguishability of inpainted regions created by diffusion models present substantial challenges, making it difficult for traditional forgery localization techniques to detect and accurately localize these tampered areas. 

\myheading{Synthetic image detection.} Early research has focused on detecting images generated by GANs, where various studies utilized hand-crafted features such as color \citep{mccloskey2018detecting-color}, saturation \citep{mccloskey2019detecting-satur}, and blending artifacts \citep{li2020face_bleding} to identify GAN-generated images. Additionally, deep learning classifiers have been employed to detect synthetic images \citep{marra2018detection}. With the rise and growing dominance of diffusion models, GANs are gradually being supplanted, and methods designed for GAN-generated images show reduced performance when applied to diffusion models. To address this, new techniques such as DIRE \citep{wang2023dire} and SeDIE \citep{ma2023exposing-sedie} leverage reconstruction error in the image space, while LaRE \citep{luo2024lare2} examines reconstruction error in the latent space for more effective detection of diffusion-generated images. Moreover, \cite{Guo2023-hifi} developed hierarchical fine-grained labeling techniques for classifying synthetic images. Recently, \citep{fakebenchLi2024} have utilized the capabilities of multimodal LLMs to study the detection of synthetic images. While the detection of generated images remains crucial for maintaining the trustworthiness of generative AI, pinpointing the exact edited regions within an image poses a more challenging and interesting problem. 


\section{\Approach}
\myheading{Problem setting:} 
Our primary objective is to accurately identify and localize regions within an image that have been altered using diffusion-based editing techniques. Given an image $\mathcal{I} \in \mathbb{R}^{H \times W \times 3}$, which is suspected to contain edited portions, our goal is to generate a binary segmentation mask $\mathcal{M} \in [0,1]^{H \times W}$ indicating edited regions. In this mask, each pixel is classified as either authentic or edited, where a value of $0$ indicates an authentic pixel, and a value of $1$ signifies an edited pixel. 

\begin{figure}
    \centering
    \includegraphics[width=1\linewidth]{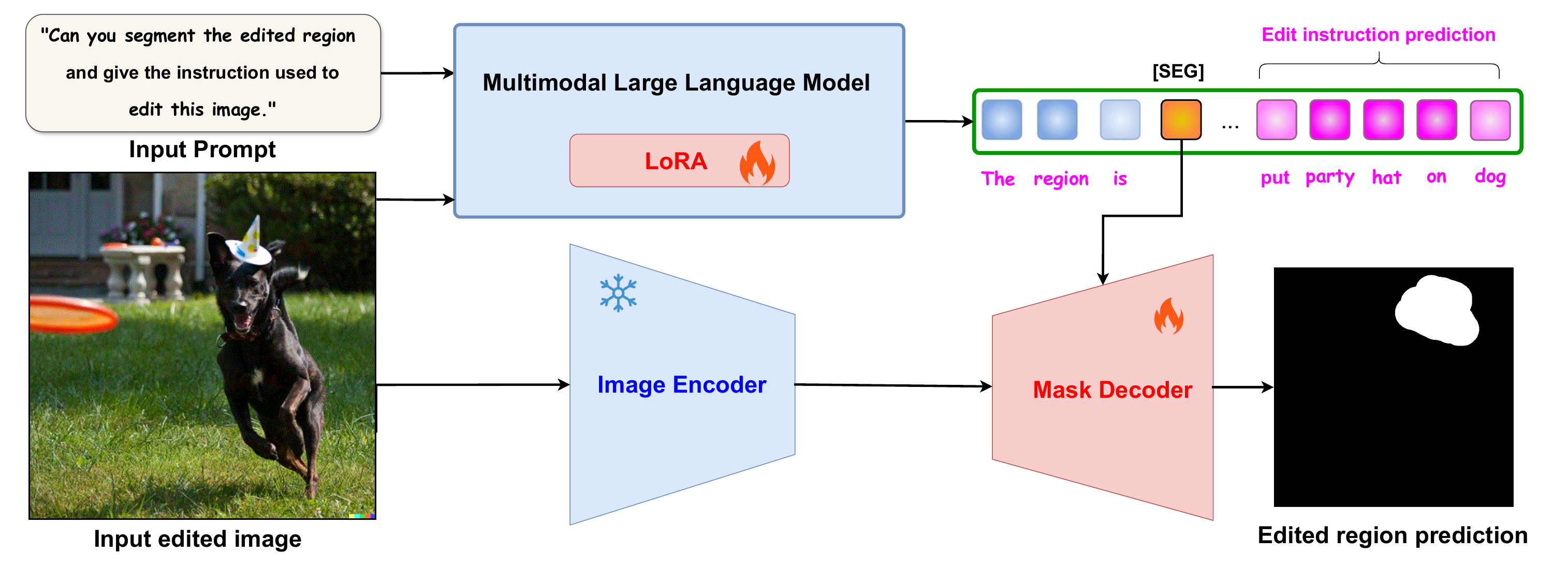}
        \caption{\textbf{Overview of \Approach}. There are two main modules: MLLM-based reasoning query generation and a segmentation model. The first module takes as input the user's prompt and image, producing a sequence of text tokens that includes a special [SEG] token representing the reasoning query and the edit instruction. The second module receives the [SEG] token as a query to generate the binary mask indicating the edited regions. Notably, only the mask decoder and a part of the MLLM are fine-tuned, while the other components remain frozen. 
        }
    \label{fig:main_diagram}
    \label{-30pt}
\end{figure}

\myheading{Overview:}
To achieve the goal, our approach is inspired by the reasoning segmentation task as described in recent works \cite{Lai2023-lisa, Yang2023-lisa++}. Specifically, we build our method, \Approach, upon the robust architectures proposed in these studies.
As shown in Fig.~\ref{fig:main_diagram}, the network consists of two primary modules: the MLLM-based reasoning query generation module and the segmentation model. The first module processes a user’s prompt along with the input image, generating a sequence of text tokens. This sequence includes a special [SEG] token that encapsulates the reasoning query and the edit instructions. The second module then uses this [SEG] token as a query to produce a binary mask that highlights the edited regions in the image.

Specifically, in the first module, given the input image $\mathcal{I}$ that appears to be edited, we also use a fixed prompt: \texttt{"Can you segment the edited region and give the instruction used to edit this image.}". Both the image and the prompt are fed into LLAVA \cite{liu2024visual}, a multimodal LLM, which produces a sequence of response tokens following this template: \texttt{"The edited region is [SEG], and the edit instruction used is $c$"}, where $c$ represents the predicted edit instruction. For instance, in Fig.~\ref{fig:main_diagram}, $c = \texttt{"put party hat on dog"}$. The [SEG] token is then converted into the reasoning query $s$ using a Multi-Layer Perceptron (MLP). At the same time, the input image is encoded by an image encoder to generate an image representation. Finally, a mask decoder uses the image representation and the reasoning query $s$ to produce the final binary mask prediction $\mathcal{M}$. It is worth noting the image encoder and mask decoder are the same as in Segment Anything Model (SAM) \cite{kirillov2023segany_SAM}.

\begin{figure}[!t]
    \centering
    \includegraphics[width=1\linewidth]{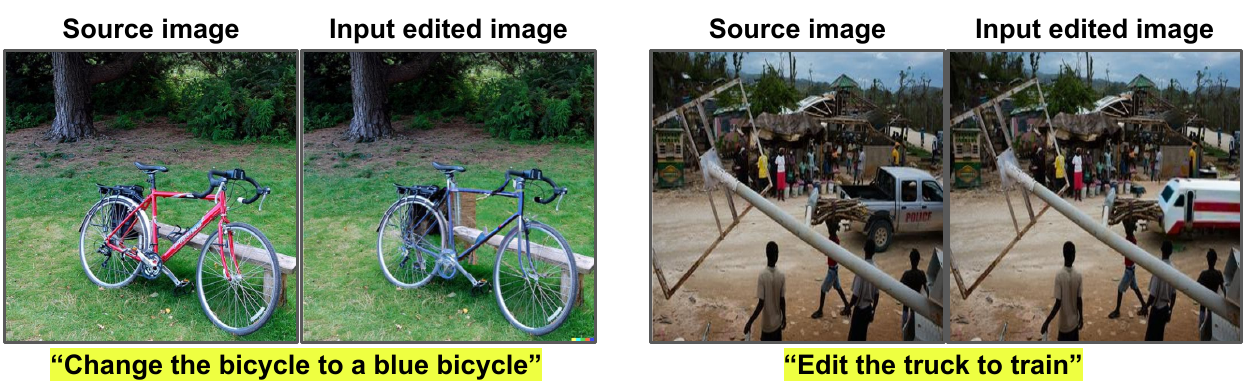}
    \caption{Edit instruction examples from the MagicBrush dataset \cite{zhang2024magicbrush} (Left) and the AutoSplice dataset \cite{jia2023autosplice} (Right).  
    }
    \label{fig:train_set}
\end{figure}

\myheading{Edit instruction construction:} 
We apply supervised learning to train \Approach~using \textbf{MagicBrush} and \textbf{AutoSplice} datasets in our experiments.
We observe that edit instructions provide strong supervision, enhancing both the reasoning capabilities of our approach and offering clear explanations for the model's predicted segmentation masks. This results in a more explainable image forensic system. To this end, we construct the ground truth for edit instructions in our training datasets.
For the \textbf{MagicBrush} dataset \cite{zhang2024magicbrush}, we utilize the provided edit instructions, which detail the specific manipulations applied to the images. However, the \textbf{AutoSplice} dataset \cite{jia2023autosplice} only provides class names for objects without detailed descriptions of the manipulations. To address this, we enhance the dataset by adding several verbs referring to manipulation activities. These verbs describe the transformation from the original object to the edited object, formatted as \texttt{<verb> <original object> to <edited object>}. For instance, an instruction might read \texttt{"replace an apple with an orange"} or \texttt{"edit dog to cat"}, providing clear guidance on the nature of the edits. We showcase several examples of our training dataset in \cref{fig:train_set}.

\myheading{Finetuning strategy:} Following the methods implemented in LISA \cite{Lai2023-lisa}, we only use LoRA \cite{hu2021lora} to finetune a part of the MLLM, fully finetune the mask decoder, and train from scratch the MLP converter. The other modules remain frozen. We use two training loss functions: (1) auto-regressive cross-entropy loss $\mathcal{L}_c$ incurred between the predicted edit instruction $c$ and the ground-truth instruction $c^*$ and (2) segmentation loss $\mathcal{L}_m$ incurred between predicted mask $\mathcal{M}$ and ground-truth mask $\mathcal{M}^*$. Particularly,
\begin{gather}
    \mathcal{L}_c = \mathcal{L}_{\text{\textbf{CE}}}(c, c^{*}), \\
    \mathcal{L}_m = \lambda_{bce}\mathcal{L}_{\text{\textbf{BCE}}}( \mathcal{M}, \mathcal{M}^{*}) + \lambda_{dice}\mathcal{L}_{\text{\textbf{DICE}}}(\mathcal{M}, \mathcal{M}^{*}), \\
    \mathcal{L} = \lambda_{c}\mathcal{L}_c + \lambda_m\mathcal{L}_m,
\end{gather}
where \(\lambda_{bce}\), \(\lambda_{dice}\), \(\lambda_{c}\), and \(\lambda_m\) are weighting parameters for the loss components, and the combination of binary cross-entroy and dice loss is common segmentation loss.

\section{Datasets}
\label{sec:dataset}

To evaluate our \Approach, we train using the training set of MagicBrush \cite{zhang2024magicbrush} and the entire AutoSplice dataset \cite{jia2023autosplice}. For testing, we use the dev and test sets of MagicBrush \cite{zhang2024magicbrush}, along with CocoGLIDE \cite{Guillaro2022-trufor}, and \Ourtestset~- \textbf{our contributed dataset}. We provide details of the composition of the training and testing datasets, including the number of samples and the methods of editing in the Tab.~\ref{tab:summarize_data}

\begin{table}[!t]
    \centering
    \caption{Summary of our training and test datasets.}
    \begin{tabular}{lcccc}
    \toprule
         \bf Datasets &\bf Train & \bf Test & \bf Editing method & \bf \#Samples  \\
    \midrule
         AutoSplice \cite{jia2023autosplice} &\checkmark  &  & DALL-E \cite{ramesh2021zero} & 3,621  \\ 
         MagicBrush \cite{zhang2024magicbrush} (\textit{train}) & \checkmark &  & DALL-E \cite{ramesh2021zero} &  4,512   \\ 
         MagicBrush \cite{zhang2024magicbrush} (\textit{dev + test})& & \checkmark  & DALL-E \citep{ramesh2021zero} & 801   \\ 
         CocoGLIDE \cite{Guillaro2022-trufor} & &\checkmark & GLIDE \citep{nichol2021glide} & 512 \\
        PerfBrush (Our proposed dataset) && \checkmark & BrushNet \citep{ju2024brushnet} & 801  \\ 
    \bottomrule
    \end{tabular}
    
    \label{tab:summarize_data}
    \vspace{-5pt}
\end{table}


The \textbf{MagicBrush} dataset, originally designed for image inpainting, has been repurposed for image forensics. It comprises three splits: train, dev, and test, containing $4,512$, $266$, and $535$ images, respectively. Each image undergoes up to three rounds of editing using DALL-E \citep{ramesh2021zero, ramesh2022hierarchical_dalle1}, but we only utilize the results from the first round as input edited images.
The \textbf{AutoSplice} dataset, another diffusion-based image forensic dataset with $3,621$ images, also employs DALL-E as the editor.
Additionally, the \textbf{CocoGLIDE} dataset consists of $512$ samples from the COCO 2017 validation set, edited using the GLIDE model \citep{nichol2021glide}.

\begin{figure}
    \centering
    \includegraphics[width=1.\linewidth]{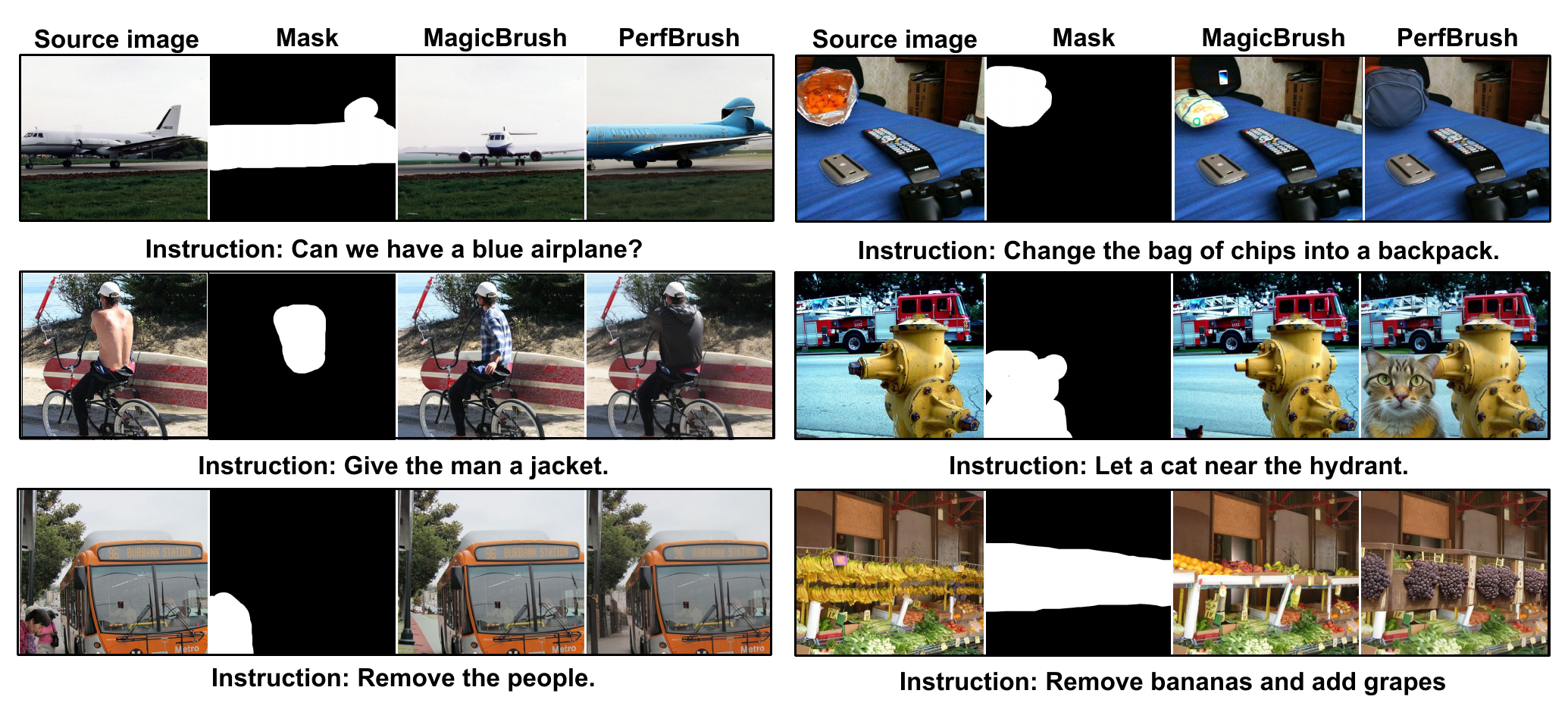}
    \caption{
        We present a series of editing comparisons between two distinct datasets: MagicBrush and PerfBrush. Leveraging the source and mask images from MagicBrush, we employ the BrushNet inpainting pipeline to generate a diversed set of edited outcomes. 
    }
    \label{fig:compare_mb_pb}
\end{figure}

Notably, to further diversify our test set with an unseen diffusion-based editing technique and to test the generalizability of image forensic models, we constructed a new set of edited images called \textbf{PerfBrush}. Instead of using DALL-E, we utilized images, instructions, and inpainting masks from the dev and test splits of MagicBrush. Specifically, we fed a triplet of a source image, a global description, and an inpainting mask to an image inpainting method called BrushNet \citep{ju2024brushnet}. To accelerate the inpainting process, we adapted the original BrushNet with PerFlow \cite{yan2024perflow}, a plug-and-play accelerator that speeds up the diffusion sampling process. For each sample triplet, we generated edited images using BrushNet with $k$ different seeds ($k=5$) to ensure diversity in the edits. We then manually selected the image with the highest quality within the inpainting region to ensure robust evaluation of image forensic methods. Our constructed test set, \Ourtestset, has better visual appearance compared to MagicBrush \cite{zhang2024magicbrush} as showcased in \cref{fig:compare_mb_pb}



\section{Experiments}
\label{sec:exp}
\vspace{-5pt}

\myheading{Evaluation metric:} We evaluate the performance of \Approach~using mean Intersection over Union (mIoU) and F1 score metrics. The mIoU (\%) score measures the overlap between the predicted binary segmentation masks and the ground truth masks for each class and takes the average across two classes. Whereas, the F1 score is calculated as the harmonic mean of the precision and recall of the pixel-level binary classification.

\myheading{Implementation details:} We build our framework on PyTorch deep learning framework \citep{paszke2017automatic_pytorch}. We utilize LoRA \citep{hu2021lora} for finetuning the LLM with lora rank is set to 8, and lora alpha is set to 32. Through experiments, we set \(\lambda_{bce} = 2.0\), \(\lambda_{dice} = 0.5\), \(\lambda_{c} = 1.0\), and \(\lambda_m=1.0\) as the best configurations.
We train \Approach~ with AdamW optimizer \citep{loshchilov2017decoupled_adamw} with a learning rate of 3$e$-5. To fairly compare with other image forensic techniques, we finetune those methods on our training dataset and follow their configurations. 
All experiments are conducted with single A100 40GB.


\subsection{Main Results}
\myheading{Quantitative results.} \cref{tab:main-results-ft} provides a detailed comparative analysis of our method, \Approach, against established forensic techniques. These methods, originally pretrained on extensive datasets featuring traditional editing techniques like copy-move or splicing, initially show poor performance across all three test sets because they cannot generalize to unseen edit techniques. However, when these models are fine-tuned with our training dataset, there is a significant improvement in their performance, particularly on the MagicBrush dataset \cite{zhang2024magicbrush}. However, this fine-tuning also tends to over-fit the models to the types of edits encountered during training. For example, while both the MagicBrush \cite{zhang2024magicbrush} and CocoGLIDE \cite{Guillaro2022-trufor} datasets, which include edits generated by DALL-E and GLIDE methods, exhibit improvements post-fine-tuning, this specialization results in a marked decrease in performance on the PerfBrush dataset. Specifically, CAT-Net \cite{Kwon2021-catnet} achieves mIoUs of 30.47 and 31.79 on the MagicBrush and CocoGLIDE datasets respectively but plummets to just 3.67 mIoU on PerfBrush, highlighting an overfitting issue. In contrast, \Approach~demonstrates good generalizability and robustness, attaining the highest mIoU of 22.55 on the challenging PerfBrush dataset, which features edits not previously encountered during training. Furthermore, our approach also outperforms other techniques on the CocoGLIDE dataset with an mIoU of 34.11
\label{sec:quant_results}.

\begin{table}[!t]
    \centering
    \small
    \caption{Comparison in mIoU and F1-score between \Approach~and other methods on MagicBrush \cite{zhang2024magicbrush}, CocoGLIDE \cite{Guillaro2022-trufor}, and PerfBrush. Note that HiFi \citep{Guo2023-hifi} and TruFor \cite{Guillaro2022-trufor} do not provide training scripts, hence, we only report the results evaluated on their provided pretrained weights. $\dagger$ denotes finetuning with our training set.}
    \begin{tabular}{lcccccc}
    \toprule
    \bf \multirow{2}{*}{Methods} & \multicolumn{2}{c}{\bf MagicBrush} \cite{zhang2024magicbrush}                    & \multicolumn{2}{c}{\bf CocoGLIDE} \cite{Guillaro2022-trufor}                     & \multicolumn{2}{c}{\bf \Ourtestset}    \\
    \cmidrule{2-7}
        & \multicolumn{1}{c}{\bf mIoU} & \multicolumn{1}{c}{\bf F1 } & \multicolumn{1}{c}{\bf mIoU} & \multicolumn{1}{c}{\bf F1 } & \multicolumn{1}{c}{\bf mIoU } & \multicolumn{1}{c}{\bf F1} \\
    \midrule
        \textbf{PSCC-Net} \cite{Liu2021-pscc} & 8.35 & 12.3 & 14.46 & 20.24 & 7.90 & 12.59\\
        \textbf{EITL-Net} \cite{eitl@2024} & 7.88 & 11.38 & 28.79 & 35.42 & 18.55 & 22.13\\
        \textbf{TruFor} \cite{Guillaro2022-trufor} &  19.47 & 26.93 & 29.26 & 36.08 & 15.66	& 22.36\\                       
        \textbf{HiFi} \cite{Guo2023-hifi} & 5.10 & 8.22 & 16.55 &	23.44 & 4.45 &	6.99\\                       
        \textbf{CAT-Net} \cite{Kwon2021-catnet} & 2.71 & 4.33 & 31.63 & 39.18 & 8.11 &	11.96\\           
    \midrule
        \textbf{PSCC-Net}$^{\dagger}$ \cite{Liu2021-pscc} & 16.82 & 26.50 & 15.02 & 20.75 & 9.71 & 15.75\\                       
        \textbf{EITL-Net}$^{\dagger}$ \cite{eitl@2024} & 20.02 & 28.09 & 19.15 & 26.34 & 8.00 & 12.16\\
        
        \textbf{CAT-Net}$^{\dagger}$  \cite{Kwon2021-catnet} &  \bf 30.47 &  \bf 40.35 &  31.79 &  41.12 & 3.67 & 5.52\\ 
    \midrule
        \bf \Approach  & 23.77 & 33.19 & \bf 34.11 & \bf 45.70 & \bf 22.55 & \bf 31.04\\
    \bottomrule
    \end{tabular}
    \label{tab:main-results-ft}
\end{table}

\begin{figure}[!ht]
    \centering
    \includegraphics[width = 1\linewidth]{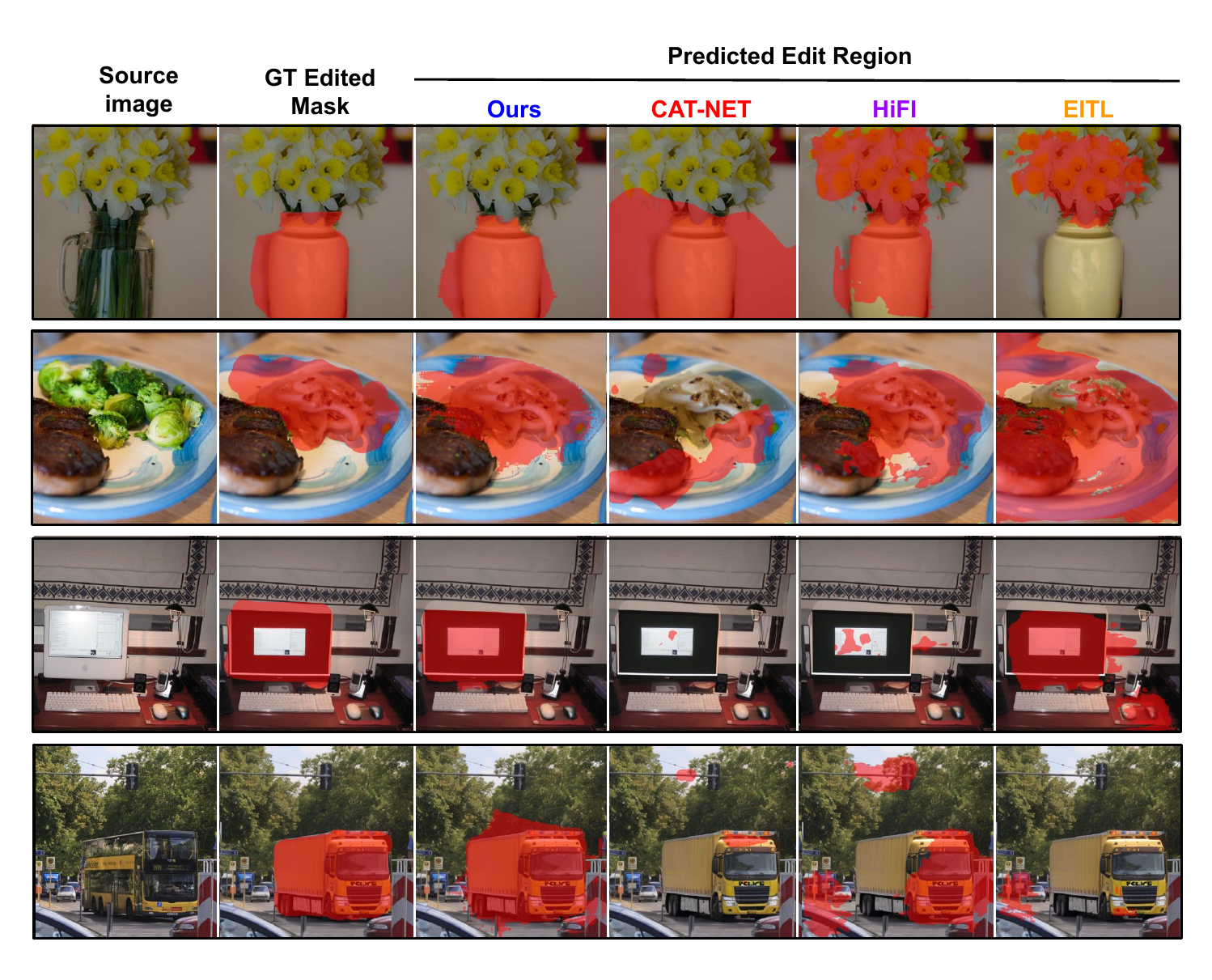}
    \vspace{-20pt}
    \caption{Comparison of predicted masks between \Approach~and other methods on the \textbf{MagicBrush} (\textit{dev} + \textit{test}) (first two rows) and \textbf{\Ourtestset} (last two rows) datasets.
    }
    \label{fig:qual_combine}
    \vspace{-10pt}
\end{figure}

\begin{figure}[!ht]
    \centering
    \includegraphics[width=\linewidth]{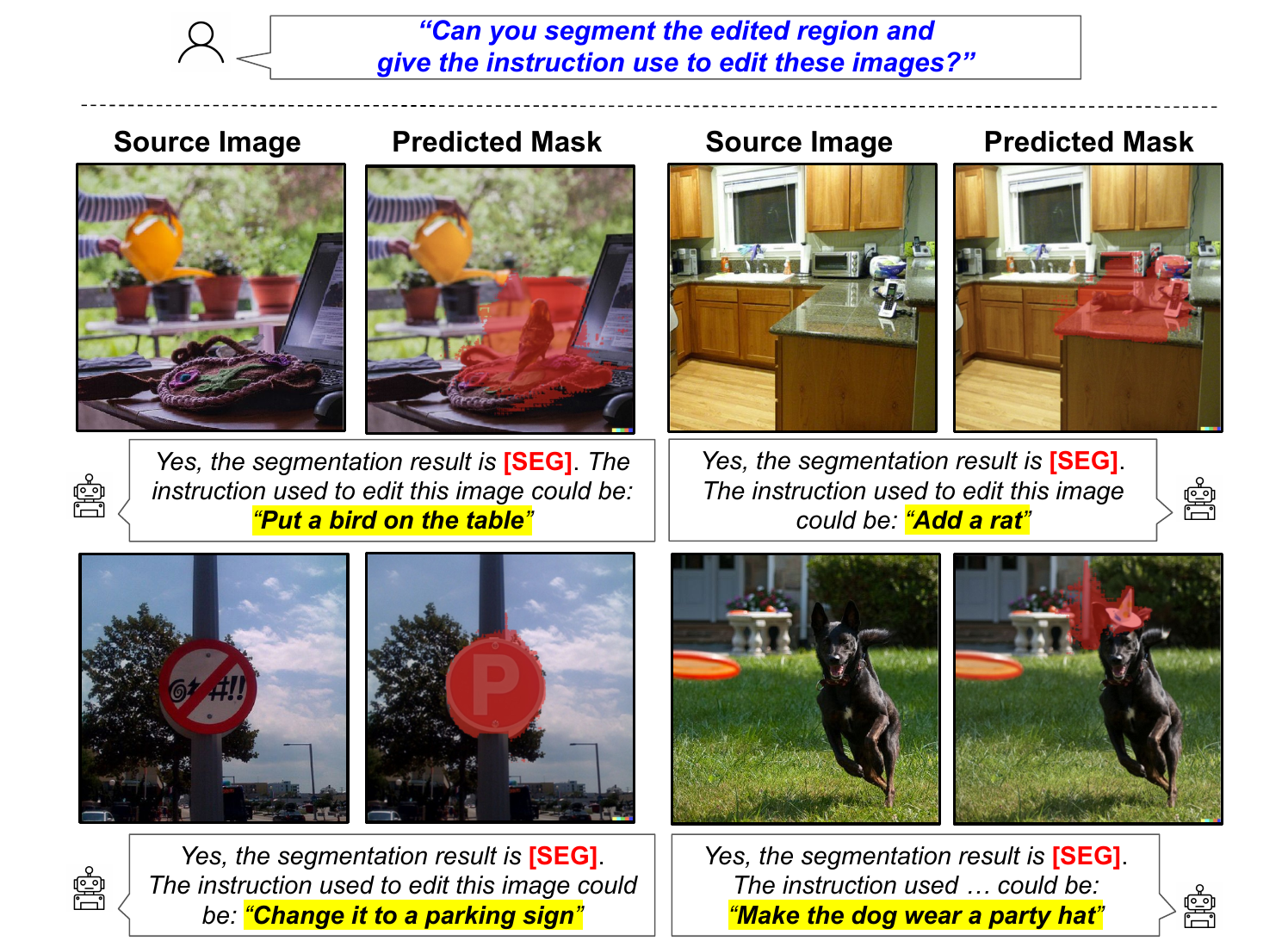}
    \caption{Illustrative visualization for the reasoning capability of \Approach~across a diverse range of samples. Users can input a natural prompt to \Approach. The results include a visual prompt depicting the predicted segmentation mask, denoted by a special \texttt{[SEG]} token, highlighting the edited region, along with a guessed edit instruction.} 
    \label{fig:qual_reason}
\end{figure}

\myheading{Qualitative results}. In \cref{fig:qual_combine}, we present visual comparison results from EditScout and several other image forensic methods evaluated on the MagicBrush (\textit{dev} + \textit{test}) and \Ourtestset~. The images illustrate how EditScout accurately identifies the edited regions, closely matching the provided ground-truth mask. In contrast, the other methods struggle to segment the correct regions, often resulting in fragmented and inaccurate prediction masks. Additionally, in \cref{fig:qual_reason}, we demonstrate the reasoning capability of EditScout. Our framework accepts natural language prompts from the user and processes them to generate comprehensive responses. These responses include both the segmentation results and predictions regarding the editing instructions. This showcases EditScout's ability not only to accurately identify edited regions but also to understand and interpret user prompts, providing insightful and detailed outputs.

\subsection{Ablation Study}
We conduct all ablation study experiments with the input prompt and training hyperparameters as discussed in implementation details.
Additionally, we report the results evaluated on the MagicBrush (\textit{dev} + \textit{test}) dataset.


\myheading{Effect of different components} on overall performance is summarized in the \cref{tab:ablate_modules}. First, we examine the performance of our method if just only use visual understanding. In setting A) using only the Segment Anything model \citep{kirillov2023segany_SAM} with a learnable \texttt{[SEG]} token in training yields a very low result which is 5.96 mIoU. In contrast, with setting B) using image embedding produced by the vision encoder of multimodal large language model as \texttt{[SEG]} significantly boosts performance to 14.94. Then, we enable the reasoning capability with LLM by using input prompt which yields a much better result of 22.23 mIoU. Finally, we obtain the best result using edit instruction with 23.77 mIoU.

\begin{table}[!t]
    \centering
    \small
    \caption{Impact of different components in overall performance}
    \label{tab:ablate_modules}
    \setlength{\tabcolsep}{5pt}
    \begin{tabular}{ccccccc}
    \toprule
     & \multicolumn{1}{c}{CLIP image embedding to LLM} & Using input prompt       & \multicolumn{1}{c}{Edit Inst. Prediction} & \textbf{mIoU}   & \multicolumn{1}{c}{\textbf{F1}} &                      \\
    \midrule
    A  & \multicolumn{1}{c}{}                     & \multicolumn{1}{c}{} & \multicolumn{1}{c}{}                        & \multicolumn{1}{c}{5.96} & \multicolumn{1}{c}{9.83}                 & \multicolumn{1}{c}{} \\ 
    B   & \checkmark                                        &                       &                                             &                      14.94&21.98                                      &                      \\
    C   & \checkmark                                        & \checkmark                    &                                             &22.23                      &31.55                                      &                      \\
    D & \checkmark                                        & \checkmark                    & \checkmark                                           &                      \bf 23.77& \bf 33.19                                      &                     \\
    \bottomrule
    \end{tabular}
\vspace{-10pt}
\end{table}

\myheading{Effect of different choices of input prompt.} We conduct sensitivity analysis on input prompt to the LLM to determine the best question template for finetuning and report the results in Tab.~\ref{tab:ablate_prompting}.  The first prompt, which explicitly asks for segmentation and an explanation of the editing technique, yields the best result of 23.77 mIoU. It suggests that direct queries may facilitate more accurate and detailed analyses by the MLLMs. The subsequent prompts show a gradual decrease in performance metrics indicating that less specific or more ambiguously phrased prompts may lead to less precise localization performance.

\begin{table}[!t]
\footnotesize
    \centering
    \small
    \caption{Effect of different prompts to query the LLM}
    \begin{tabular}{p{11cm}cc}
    \toprule
    \textbf{Input Prompt} & \bf mIoU & \bf F1 \\ 
    \midrule
          ``Can you segment the edited region and give the instruction used to edit this image.'' &  \bf 23.77 & \bf 33.19\\
    \midrule
         ``Could you segment the modified regions and provide a detailed explanation of the editing process?''&  20.20 & 28.69\\
    \midrule 
         ``Please analyze this image for any signs of editing. If the image has been edited, identify and segment the edited portions, and outline the steps taken to achieve the edits.'' &  19.90 & 28.35\\
    \midrule
          ``Can you determine if this image has been manipulated? If so, please highlight the altered areas and describe the techniques used to modify the image.'' &  19.27 & 27.46\\
    \midrule 
    Randomly choose one of the above input prompts & 19.32   & 31.54  \\ 
    \bottomrule
    \end{tabular}
    
    \label{tab:ablate_prompting}
\end{table}

\section{Discussion and Conclusion}
\label{sec:conclusion}
\myheading{Limitations:} 
Firstly, the binary masks produced by our approach are not perfect. We expect integrating modules specially designed for image forensics can improve the segmentation capability. Secondly, while our method can identify potential editing activities, it does not provide explanations for why specific edits were made. Future research direction can develop datasets and techniques that facilitate reasoning about image manipulations. 

\myheading{Conclusion:} We introduce a novel approach, \Approach, for detecting forgeries in diffusion-based edited images. Utilizing the extensive visual and linguistic knowledge of MLLMs, \Approach effectively localizes subtle signs of tampering in the content. Our method outperforms existing ones, achieving 23.77 mIoU on MagicBrush, 34.11 mIoU on CocoGLIDE, and 22.55 mIoU on \Ourtestset. These results confirm our approach's potential and suggest promising avenues for further integrating foundation models to enhance digital image forensics and the credibility of generative AI.

{
    \bibliographystyle{unsrt}
    \bibliography{main}
}


\end{document}